\documentclass{MechatronikTagung}

\usepackage[utf8]{inputenc} 
\usepackage[T1]{fontenc}

\usepackage{amsmath} 
\usepackage{amssymb} 

\usepackage{graphicx}

\usepackage{booktabs} 
\usepackage{cite} 

\usepackage{wrapfig}
\usepackage{blindtext}
\usepackage{array}   
\newcolumntype{C}[1]{>{\centering}p{#1}}
\usepackage[justification=centering]{caption}
\usepackage{subcaption}
\usepackage{siunitx}
\DeclareMathAlphabet{\mathcal}{OMS}{cmsy}{m}{n}
\DeclareSymbolFont{newfont}{OML}{cmm}{m}{it}
\DeclareMathSymbol{\Epsilon}{3}{newfont}{15}
\usepackage{units}
\usepackage{hyperref}
\newcommand{\cmmnt}[1]{}

\titelenglisch{Autoencoder-based Representation Learning from Heterogeneous Multivariate Time Series Data of Mechatronic Systems}

\autorenliste{Karl-Philipp Kortmann, M.\,Sc.$^{\dagger \ddagger}$, Moritz Fehsenfeld, M.\,Sc.$^{\dagger}$ and Dr.-Ing. Mark Wielitzka$^{\dagger}$ \\[6pt]
$^{\dagger}$\;Leibniz University Hannover, Institute of Mechatronic Systems, An der Universität 1, 30823 Garbsen, Germany \\
$^{\ddagger}$\;kortmann@imes.uni-hannover.de
}

 \kurzfassungenglisch{Sensor and control data of modern mechatronic systems are often available as heterogeneous time series with different sampling rates and value ranges. Suitable classification and regression methods from the field of supervised machine learning already exist for predictive tasks, for example in the context of condition monitoring, but their performance scales strongly with the number of labeled training data. Their provision is often associated with high effort in the form of person-hours or additional sensors. In this paper, we present a method for unsupervised feature extraction using autoencoder networks that specifically addresses the heterogeneous nature of the database and reduces the amount of labeled training data required compared to existing methods. Three public datasets of mechatronic systems from different application domains are used to validate the results.}

\begin{document}

\section{Introduction}
Modern mechatronic systems contain a large number of internal sensor and control signals that can be accessed for example via fieldbus interfaces. In addition, these systems can be extended by external measuring systems, which in total leads to a heterogeneous set of multivariate time series signals (MTS), where heterogeneous here implies divergent sampling times, measuring resolutions or scale levels of the individual univariate signals.\\
In the course of progressive digitization, such system signals are increasingly used for classification tasks (such as \textit{alarm} or \textit{condition monitoring}) or for the regression of target variables that cannot be measured directly (e.g. process stability or quality). In the following, these are collectively referred to as \textit{estimation tasks}.\\
The estimation methods primarily used for this purpose from the field of statistical or machine learning depend on extensive feature extraction \cite{fawaz2019}, especially in the case of a high-dimensional, heterogeneous data basis. In the case of manual feature extraction, this requires a high level of technical expertise regarding the system at hand and is consequently accompanied by a high effort during implementation. Alternatively, supervised end-to-end estimation methods from the field of deep learning explicitly manage without previously extracted features, but require a large amount of already labeled data for training \cite{lei2016}.\\
Methods of \textit{representation learning} have recently emerged in the field of computer vision and speech recognition, enabling unsupervised feature extraction that outperforms the prediction performance of common manual and automated feature extraction methods in the case of only a small amount of existing labeled data \cite{Franceschi2020}. The application to mechatronic systems has so far been limited to specialized single solutions \cite{Chen.2020}, which cannot be directly transferred to any arbitrary mechanical or electrical system.

\section{State of Research}
\subsection{Representation Learning}
\begin{figure}[htbp]
  \centering
  \includegraphics[width=0.90\columnwidth, trim={0cm 0cm 0cm 0cm}]{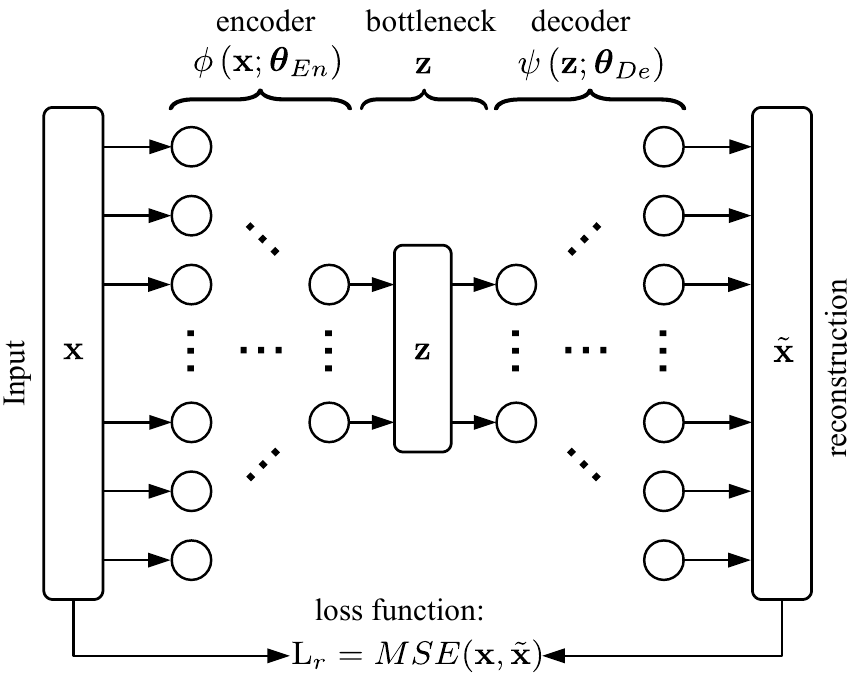}\vspace{-3.9mm}
  \caption{Schematic network diagram of a simple autoencoder with one-dimensional input and implied fully connected layers in encoder and decoder}
  \label{fig:ae} 
\end{figure}

\begin{figure}[htbp] 
  \begin{subfigure}[b]{0.5\linewidth}
    \centering
    \includegraphics[width=0.85\linewidth]{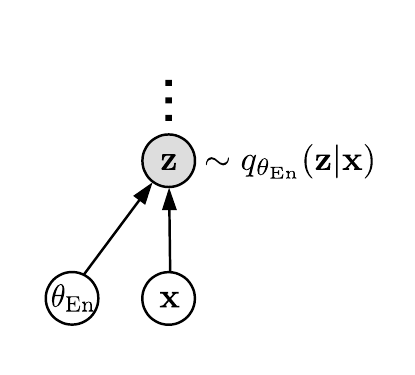} 
    \caption{} 
    \label{fig:repa_a} 
    \vspace{0.3cm}
  \end{subfigure}
  \begin{subfigure}[b]{0.5\linewidth}
    \centering
    \includegraphics[width=0.85\linewidth]{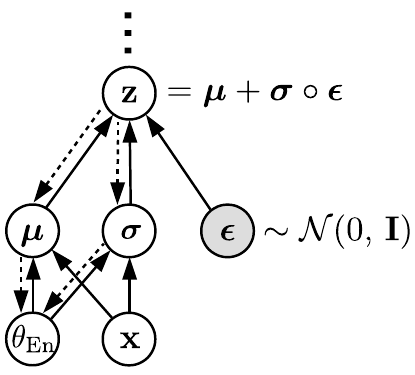} 
    \caption{} 
    \label{fig:repa_b} 
    \vspace{0.3cm}
  \end{subfigure}\vspace{-5.5mm}
  \caption{\textbf{(a)}: Direct sampling of $\mathbf{z}$ makes backpropagation ($\dashrightarrow$) of the loss function impossible. \textbf{(b)}: Only with the reparametrization of $\mathbf{z}$\cmmnt{by swapping into a normally distributed random variable $\mathbf{\Epsilon}$} the optimization of the encoder parameters becomes possible. Independent random variables are shaded gray. Based on \cite{kingma2013auto}.}
  \label{fig:repa} 
\end{figure}

The term representation learning or feature learning covers methods that allow an automatic extraction of relevant features and thus make the step of manual feature engineering superfluous in principle. In the context of this work, the focus is on an unsupervised method that learns a compact representation of the MTS without knowledge of the target variable and thus only on the basis of the input time series.
Originally derived from the field of computer vision, feature learning methods are increasingly adapted for estimation tasks based on time series data such as Chen et al. \cite{Chen.2020} using an autoencoder for feature extraction from the torque signals of a 6-axis industrial robot to predict collisions in the workspace. Li et al. \cite{li2019} and Jiang et al. \cite{jiang2019} both use a \textit{Generative Adverssarial Network} (\textit{GAN}) for feature extraction from industrial time series data, while Franceschi et al. \cite{Franceschi2020} use a pure encoder-based network in combination with a so-called \textit{triplet loss function} for classification on various reference time series.\
In this work, we focus on an autoencoder-based approach because, in contrast to GANs for instance, these generally provide a better representation of the population of training data, often at the cost of worse performance than purely generative models \cite{grover2018} (i.e., in generating realistic new time series), but this is not the focus of this paper.
\paragraph{Autoencoder}
An autoencoder is an artificial neural network whose primary goal is to reconstruct an input signal $\textbf{x}$ (see Figure~\ref{fig:ae}). The dimension-reduced latent variable (also \textit{bottelneck} or \textit{latent space})
\begin{equation}
\textbf{z}=\phi\left(\textbf{x}; \boldsymbol{\theta}_{En} \right)
\end{equation}
is the result of an encoder function with the parameters (weights) $\boldsymbol{\theta}_\mathrm{En}$ and is used as a feature vector in the representation learning in this work. The latent variable subsequently passes through a decoder 
\begin{equation}
\tilde{\textbf{x}}=\psi\left(\textbf{z}; \boldsymbol{\theta}_{De} \right)
\end{equation}
with parameters $\boldsymbol{\theta}_{\text{De}}$, which generates a reconstruction $\tilde{\mathbf{x}}$ of the input signal.
In the present case of mostly real-valued input values, usually the mean squared error ${L_\mathrm{r}= \text{MSE}(\textbf{x}, \tilde{\textbf{x}})}$
is used as reconstruction error and serves as a loss function during optimization.\\
Besides fully connected or dense layers with nonlinear activation functions, more complex neural layers also find use. For example, Bianchi et al. use recurrent (\textit{RNN}) bidirectional layers in combination with an additional kernel loss function for feature extraction from MTS with missing values \cite{bianchi2019}.
\paragraph{Variational Autoencoder}
An extension of the regular autoencoder is the \textit{Variational Autoencoder (VAE)}, which is mostly used as generative model in the field of computer vision. As an example of a \textit{Variational Bayes} model, the VAE models the unknown distribution function of the input data $\textbf{x}\sim p^{\ast}(\textbf{x})$ using a model distribution {${p_{\theta}(\textbf{x})\approx p^{\ast}(\textbf{x})}$} \cite{kingma2019}. The stochastic decoder can therefore be understood as a conditional probability distribution $p_{\theta_{\text{De}}}(\textbf{x}| \textbf{z})$, which together with the prior distribution of the latent variable $p_{\theta}(\textbf{z})$ forms a generative model by factorizing the multivariate distribution
\begin{equation}
    p_{\theta}(\textbf{x}, \textbf{z})=p_{\theta}(\textbf{z})p_{\theta_{\mathrm{De}}}(\textbf{x}|\textbf{z}).
\end{equation}
Similarly, the encoder represents an inference model that can be conceived as a conditional probability distribution of the latent variables given input data $q_{\theta_{\text{En}}}(\textbf{z}|\textbf{x})$.
This approach leads by the application of the \textit{evidence lower bound, ELBO} to the modified loss function \cite{kingma2019}
\begin{multline}
    L_{\theta_{\text{En}},\theta_{\text{De}}}(\mathbf{x}) = D_{\mathrm{KL}}(q_{\theta_{\text{En}}}(\mathbf{z}|\mathbf{x})\Vert p_{\theta}(\mathbf{z})) \\
    -\mathbb{E}_{q_{\theta_{\text{En}}}(\mathbf{z}|\mathbf{x})}\big(\log p_{{\theta}_\mathrm{De}}(\mathbf{x}|\mathbf{z})\big),
\end{multline}
with the Kullback-Leibler divergence $D_{\mathrm{KL}}$, which penalizes the deviation between a given prior distribution of the latent variable $\mathbf{z}$ and its actual (empirical) distribution given by the encoder. The second term represents the reconstruction error. The standard multivariate normal distribution $p_{\theta}(\textbf{z}) \sim \mathcal{N}(\mathbf{0},\,\mathbf{I})$ is mostly used as prior for the latent variable.
For the training using the standard stochastic gradient descent (\textit{stochastic gradient descent, SGD}) method, the gradient of the loss function $\nabla L_{\theta_{\text{En}}}, \theta_{\text{De}}(\mathbf{x}_{\mathrm{mb}})$ is calculated for each mini-batch of training data $\mathbf{x}_{\mathrm{mb}}$ in order to perform minimization of the loss function as a function of the network parameters $\theta_{\text{En}}$ and $\theta_{\text{De}}$ (\textit{backpropagation}). As can be seen in Figure \ref{fig:repa_a}, this is not possible when directly drawing $\mathbf{z} \sim q_{\theta_{\mathrm{En}}}(\mathbf{z|x})$, since the backpropagation is interrupted by the random variable~$\mathbf{z}$ \cite{rezende2014}. Only with a mathematically equivalent reparametrization by swapping out the random variation into the random variable~$\mathbf{\Epsilon}$ (which is typically modeled as normally distributed), a backpropagation of the error through the encoder is possible (so-called \textit{reparameterization trick}, see Figure \ref{fig:repa_b}).

\section{Methodology}
This section presents the suggested autoencoder-based model, as well as the embedding pipeline. This is followed by details on the experimental design, in particular the chosen comparison methods and the selected publicly available datasets.
\subsection{Feature Extraction}
Unsupervised feature extraction from MTS with the amount of univariate signals $n_{\mathrm{sig}}$ is performed using multiple VAEs trained separately for each univariate time series. This, in contrast to common two-dimensional convolutional network architectures (\textit{convolutional neural networks, CNN}), allows a parallelization of the training and an individual architecture of the networks, depending on the sampling rate of the heterogeneous signals\footnote{On the other hand, existing cross-correlations between the individual univariate time series are no longer directly observable.}.\\
As an aggregated feature, the $1\mathrm{D}$ concatenation of the individual latent variables ${\mathbf{x}_i,\;i \in \{1,...,n_{\mathrm{sig}}\}}$ is used as input of the estimator. Table~\ref{tbl:vae} lists the most important {(hyper) parameters} of the trained model. These were kept constant across all datasets in order to provide the best possible demonstration of generality.\\
\begin{table}[hbt!]
    \centering
    \caption{Summary of the most important (hyper-)parameters of the univariate VAE model.}
    \label{tbl:vae}
    \footnotesize
    \vspace{6mm}
        \begin{tabular}{p{2.3cm}cp{3.5cm}}
        \hline 
        (Hyper-)parameter & Value & Remark \\ \hline
        Input dimension & $1\times\ast$ & $\ast$: Maximum window length \\
        Compression ratio $\kappa$ & $25$ & Quotient of $\ast$ and $\mathrm{dim(\mathbf{z})}$ \\
        Hidden layer & $[\nicefrac{\ast}{2},\,\nicefrac{\ast}{2}]$ & Two hidden layers \\
        Activation radio & - & $\mathrm{tanh}$ resp. lin. for output layer \\
        Regularization & - & Early stopping and $\mathrm{L}_2$ norm \\
        Optimizer & Adam & SGD optimizer \cite{kingma2014} \\
        Batch-size & $64-512$ & $\sim n_{\mathrm{train}}$ \\
        Normalization & - & Instance or layer norm. \\
        Learning rate & \num{1e-4} & No learning rate scheduler \\
        Max. epochs & \num{1e3} & Note: early stopping \\\hline
    \end{tabular}
\end{table}
Instead of specifying a constant dimension of the latent space, a constant compression rate~$\kappa$ is chosen so that the number of extracted features scales proportional to the sampling rate and signal duration. Furthermore, during training and inference, \textit{instance normalization} is performed for each windowed input time series to account for divergent ranges of signal values and stabilize convergence during the training.
\subsection{Pipeline}
\begin{figure}[ht]
  \centering
  \includegraphics[width=0.95\columnwidth]{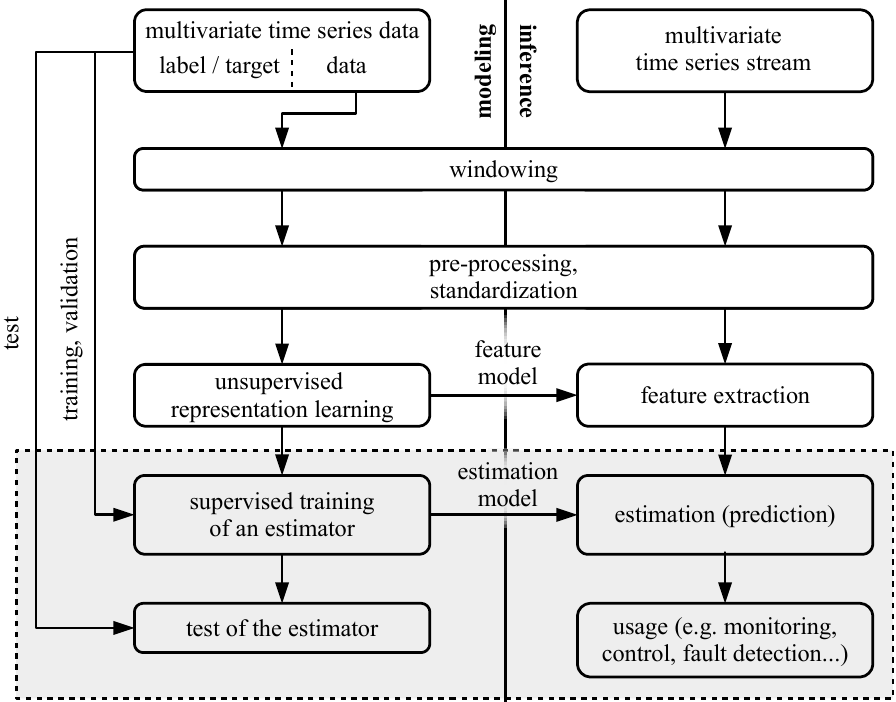}
  \caption{Flowchart of the semi-supervised ML pipeline. The consecutive estimation task is highlighted in gray.}
  \label{fig:stable-model}
\end{figure}

The sequence of modeling/training and inference steps is typical for a so-called \textit{semi-supervised} training procedure, consisting of unsupervised representation learning and supervised training of an estimator. (see Figure \ref{fig:stable-model}):
After consecutive training of both models, they are applied unchanged during inference ( in this case on independent test datasets).
The amount of labeled training data is varied during the testing procedure (logarithmic scaling of quantity).\\ The feature learning method (VAE) is provided with all available training data without labels beforehand in order to learn representations. This corresponds to the realistic use case of having a large amount of raw data, but only a certain fraction of it has been labeled. For each dataset, method, and quantity of labeled training data, $n_{\mathrm{repeat}} = 10$ replicates are performed so that empirical mean standard deviation of the particular performance metric can be calculated. The train/test split of the datasets was provided by the authors of the same.\\
Python~3.8\footnote{In particular: torch~1.7.1, sktime~0.4.3 and sklearn~0.24.1.} on a computer with GPU support (CUDA~7.5) is used as experimental environment. The source code and links to the used datasets are made publicly available\footnote{\url{https://github.com/MrPr3ntice/vae_rep_learn_mts}}.
\subsection{Comparison methods}
\begin{table*}[ht!]
    \centering
    \caption{Overview of selected data sets; $n_\mathrm{max}$ indicates the maximum number of data points in a time series sample (windowed).}
    \label{tbl:beispieltabelle}
    \footnotesize
    \vspace{4mm}
        \begin{tabular}{p{2.4cm}C{1.2cm}C{1.2cm}C{1.5cm}C{1.9cm}p{3.9cm}c}
        \hline 
        \multicolumn{1}{c}{{Dataset}} &
          {Train samples} &
          {Test samples} &
          {Number of channels} &
          {Duration per sample ($n_{\mathbf{max}}$)} &
           \multicolumn{1}{c}{{Target value (type)}} &
          {Reference} \\ \hline
        Rolling\;bearing\;damages &
          1440 &
          800 &
          7&
          $\SI{0,2}{s}\,(800)$ &
          Rolling\;bearing\;condition\;(3\;classes) &
          \cite{Lessmeier2016} \\
        Stepper motors &
          70152 &
          23384 &
          7&
          $\SI{6}{ms}\,(60)$ &
          Operating\;condition\;(4\;classes) &
          \cite{Goubeaud2020} \\
        Hydraulic system &
          1544 &
          661 &
          17&
          $\SI{60}{s}\,(1200)$  &
          Cooler condition\;(3\;classes), Hydraulic\;accumulator\;pressure\;(regr.) &
          \cite{Helwig2015} \\ \hline
        \end{tabular}
    \end{table*}
The presented representation learning model (VAE) as well as the semi-supervised pipeline are compared with a feature extraction method from the current state of the art (\textit{Rocket}). Additionally, principal component analysis serves as a deterministic baseline comparison method. The Python library \textit{tsfresh} allows automatic extraction and selection of statistical time and frequency domain features, thus providing a comparison method from the field of manual feature extraction.\\
For all comparison methods, a ridge regression-based estimator is used in that parameterization, as it is recommended as a favored predictor, especially by authors of several current feature extractors \cite{LeNguyen2019, Dempster2020}. In case of the VAE, a \textit{support vector machine (SVM)} with Gaussian kernel is used to better account for the forced normal distribution character of the latent variable\footnote{In particular, due to the symmetric kernel, closed intervals on one variable can be separated in case of classification.}.
\paragraph{PCA}
Using principal component analysis (PCA), time series data can be reduced in dimension by the help of singular value decomposition. PCA produces those orthogonal linear transformations of the input data which maximize the variance of the components of the target subspace in descending order. The obtained transformation can be used as a feature for classification or regression. In particular, PCA can be considered as a special case of a linear autoencoder without any hidden layers \cite{kingma2019}, which is why it will serve as a comparative baseline method here.
\paragraph{Statistical Features}
The extraction of manually or automatically selected statistical features from time series for ML applications is widely used.  The Python library \textit{tsfresh} provides a collection of established extraction methods to automatically generate and select a variety of these features from time series data. The collection of features includes, for example, statistical ratios and correlations in both time and frequency domain. A complete overview of the extracted features can be taken from the libraries documentation \cite{Christ2018}. In the present case, the default settings\footnote{extraction: \texttt{efficient}, selection: \texttt{extract\textunderscore relevant\textunderscore feat} \texttt{ures()}} is used.
\paragraph{Rocket}
\textit{\textbf{R}and\textbf{o}m \textbf{c}onvolutional \textbf{ke}rnel \textbf{t}ransform} is a state-of-the-art method that uses randomly sampled convolution kernels to extract features from time series. Subsequently, a Ridge regression\footnote{A special case of Tikhonov regularization for linear regression, sometimes also referred to as $L_2$-regularization \cite{Kennedy2003}. It can be used in both classification and regression case.} is trained with the thereby generated features and the known target values. Due to this straightforward setup, the computational cost of \textit{Rocket} is lower than that of comparably performing methods \cite{Dempster2020}. To the state of this work, the method produces the best average classification results on the datasets of the UCR and UEA time series archive \cite{ruiz2020}.
\subsection{Data sets}
The number of publicly available data sets is limited, especially in the area of mechanical and electronic systems. For validation of the proposed method, data sets covering a wide range of mechatronic applications are selected from those available. All data sets consist of real measurement data from several sensors, which differ e.g. in sampling rates. This results in three multivariate data sets from heterogeneous time series, from which three classification tasks and one regression task are derived. An overview of the selected data sets can be found in Table \ref{tbl:beispieltabelle}.
\paragraph{Rolling bearing damages}
A widespread use case for machine learning applications is the detection of different rolling bearing damages. A comprehensive reference data set representing this use case has been published recently by Leissmeier et al. \cite{Lessmeier2016}. In addition to high-frequency sampled measurements of motor currents and housing vibration ($f_\mathrm{s} = \SI{64}{kHz}$), additional, lower-frequency data such as radial force ($f_\mathrm{s} = \SI{4}{kHz}$) and temperature ($f_\mathrm{s} = \SI{1}{Hz}$) are available for damage classification. The dataset includes different damage types of rolling bearings on the outer and inner ring as well as the data from undamaged bearings under different operating conditions. Further information on the data set can be found in \cite{Lessmeier2016}.
\paragraph{Stepper motors}
For stepper motor monitoring, there is a dataset published by Goubeaud et al.\,\cite{Goubeaud2020}. This includes measurements of current, voltage, and vibration (translational acceleration). The target variable is the operating mode of the stepper motor, which differentiates between clockwise and counterclockwise operation, as well as operation in the normal range and beyond the mechanical stop.  A detailed description of the experimental setup and the data acquisition is given in the original publication \cite{Goubeaud2020}.
\paragraph{Hydraulic system}
The hydraulic system described by Helwig et al.\,\cite{Helwig2015} is equipped with a variety of different sensors. In addition to measurements of pressure and flow rates also temperature, current, and vibration are recorded.The sampling rates range from $f_\mathrm{s} = \SI{1}{Hz}$ (temperature) to $f_\mathrm{s} = \SI{20}{Hz}$\footnote{Compared to the original data set sampled at $\SI{100}{Hz}$, pressure has been sampled down to $\SI{20}{Hz}$ in favor of the computation time.} (pressure), resulting in a heterogeneous data set with a wide variety of sensor types, value range and sampling rate. In this work, the state of the cooler (fault classification) and the pressure in the hydraulic accumulator (regression) will be considered as target variables.

\section{Results}
\begin{figure*}[!hbt] 
  \begin{subfigure}[b]{0.5\linewidth}
    \centering
    \includegraphics[width=0.90\linewidth, trim={0cm 0.1cm 0cm 0.65cm}, clip]{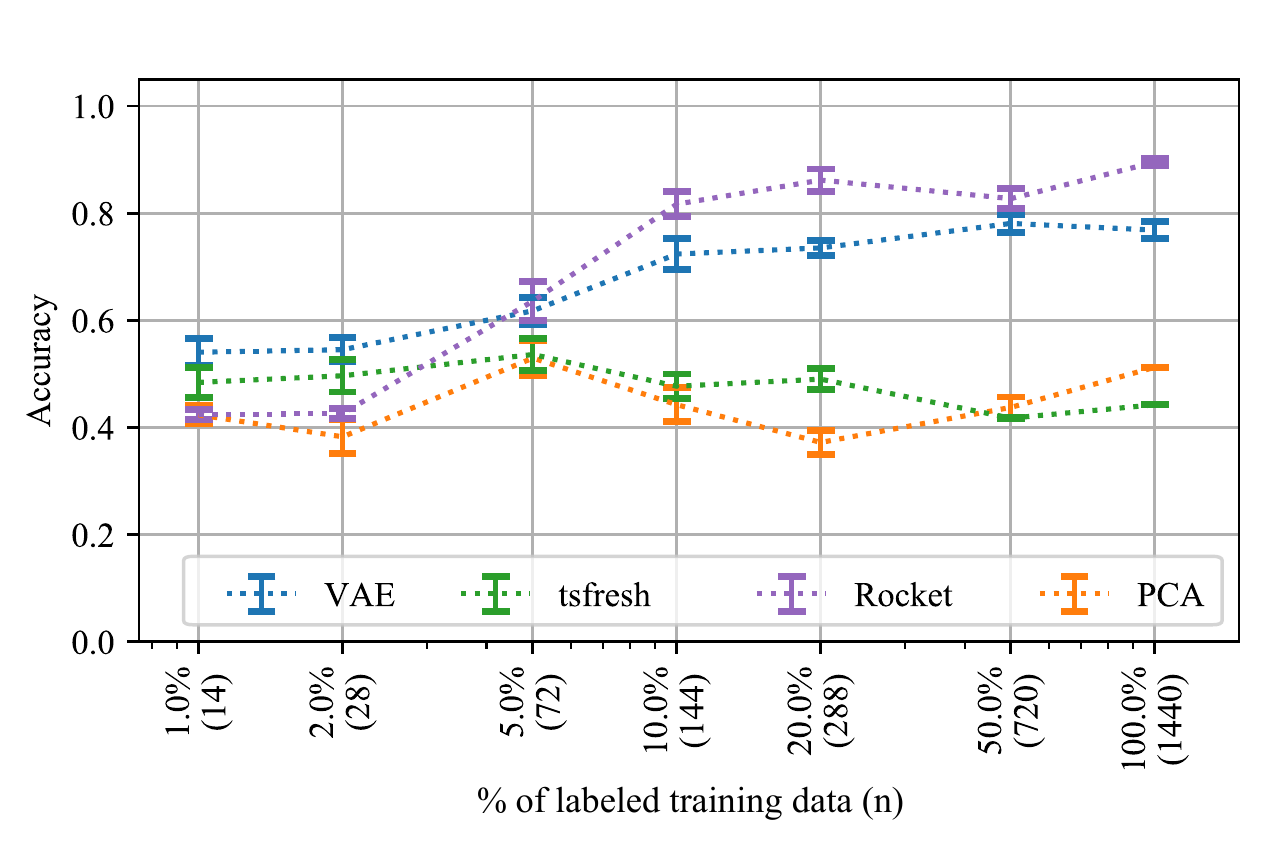}
    \vspace{-0.27cm}
    \caption{Rolling bearing damages} 
    \label{fig7:a}
  \end{subfigure}
  \begin{subfigure}[b]{0.5\linewidth}
    \centering
    \includegraphics[width=0.90\linewidth, trim={0cm 0.1cm 0cm 0.65cm}, clip]{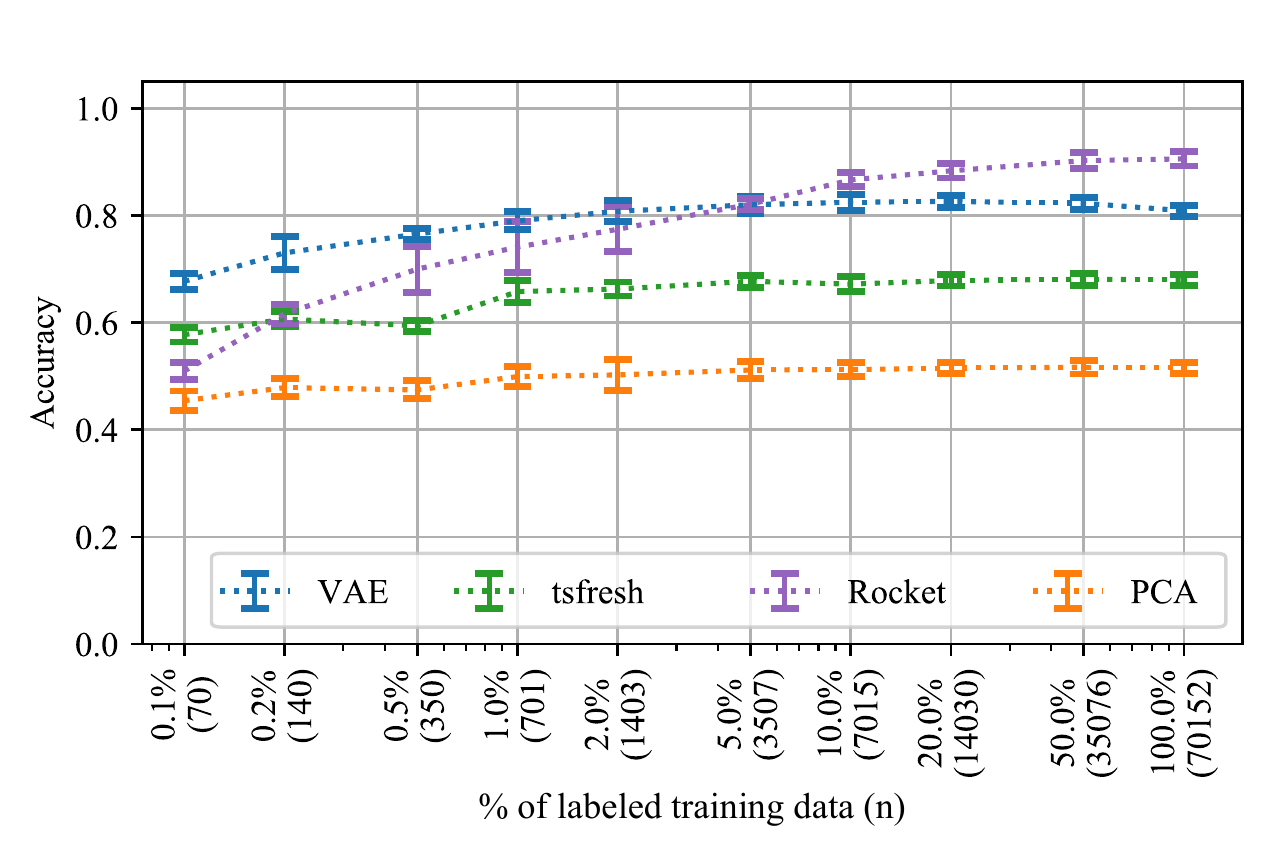}
    \vspace{-0.25cm}
    \caption{Stepper motors} 
    \label{fig7:b}
  \end{subfigure}\vspace{0.0cm}
  \begin{subfigure}[b]{0.5\linewidth}
    \centering
    \includegraphics[width=0.90\linewidth, trim={0cm 0.1cm 0cm 0.65cm}, clip]{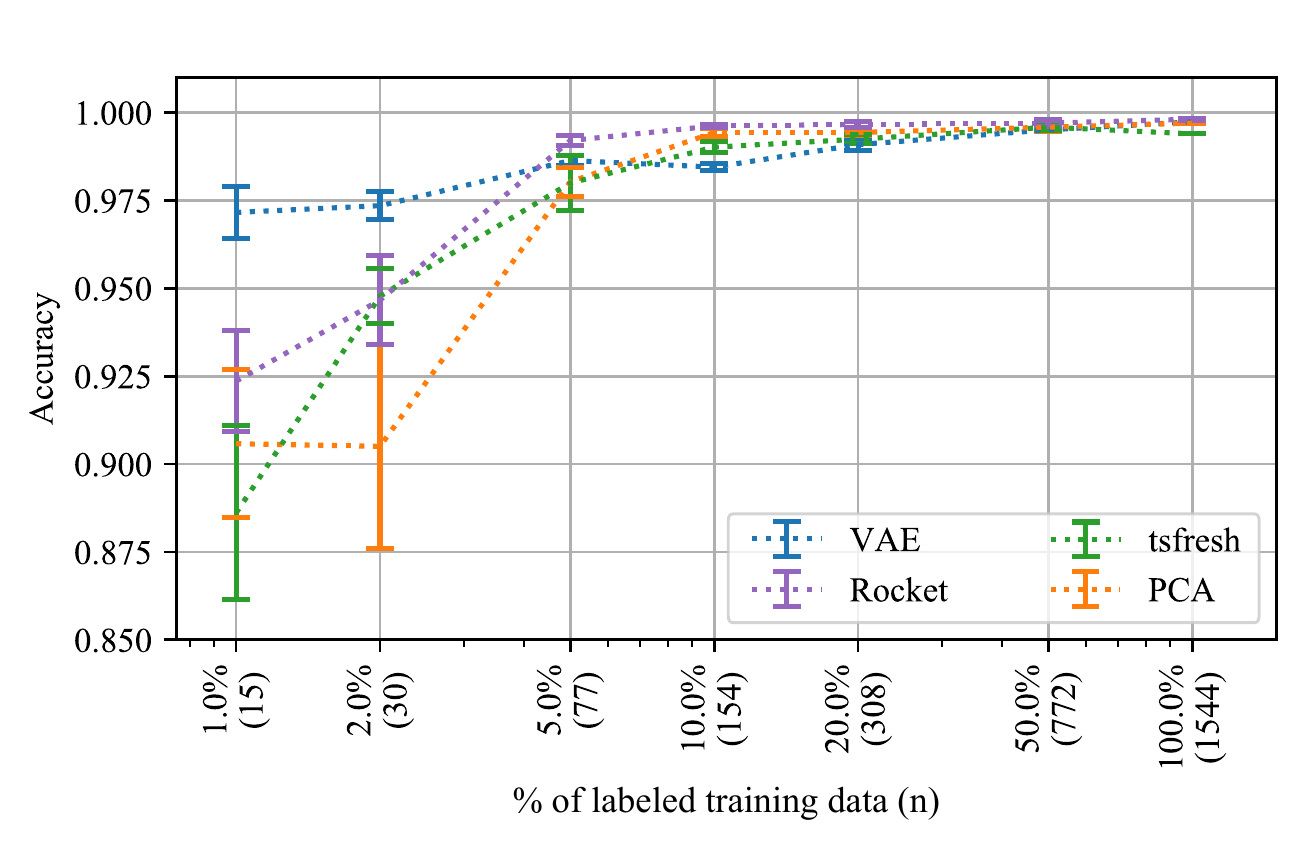} 
    \vspace{-0.25cm}
    \caption{Hydraulic system (classification)} 
    \label{fig7:c} 
  \end{subfigure}
  \begin{subfigure}[b]{0.5\linewidth}
    \centering
    \includegraphics[width=0.93\linewidth, trim={0cm 0.1cm 0.6cm 0.74cm}, clip]{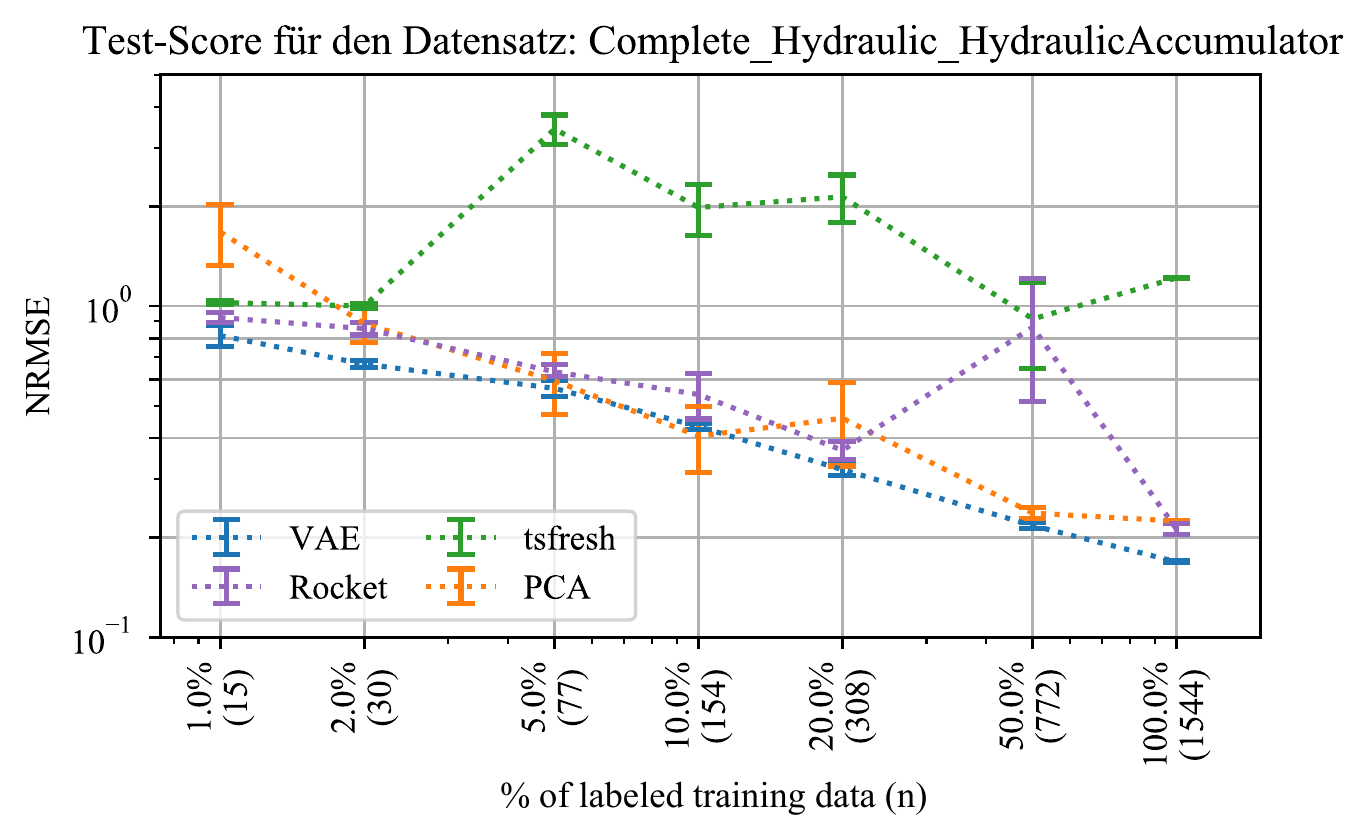} 
    \vspace{-0.22cm}
    \caption{Hydraulic system (regression)}
    \label{fig7:d} 
  \end{subfigure}\vspace{-0.40cm}
  \caption{Results of the four compared feature extractors on the three test datasets for varying proportions of labeled training data (log. scaling).The plots \textbf{(a)\,-\,(c)} show the accuracy for the respective classification tasks; \textbf{(d)} shows the normalized RMSE (log. scaling) for the regression task. For each proportion the empirical mean and standard deviation for $n_{\mathrm{repeat}}=10$ of the respective metric is shown.}
  \label{fig7} 
\end{figure*}
The results are shown in Figure~\ref{fig7} (\textbf{a})\,-\,(\textbf{d}). More detailed results can be retrieved from the Tables~\ref{tbl:results_klass} and \ref{tbl:results_regr} in the appendix.\\
For the first three cases, the \textit{Rocket} comparison method achieves the highest test results with respect to the mean accuracy. Only in the case of the pressure estimation in the hydraulic accumulator, VAE shows almost consistently the lowest RMSE. In this regard, it should be noted that \textit{Rocket} was originally developed for classification tasks and has been applied for this task in majority. Principal component analysis and the statistical features determined by \textit{tsfresh} are not competitive for the prediction tasks on the first two data sets; for classification on the hydraulic data, all methods achieve similarly low error rates for a higher fraction of labeled training data.\\
A direct comparison with the partly available results of the original publications of the data sets is not possible at this point, since the multivariate case of a training data with a varying number of labeled data, which is treated here, was not considered in most studies. Additionally a neutral selection and parameterization of the compared estimators has been used in this work, in order to ensure the best possible comparability between the methods and not the highest possible performance of the same.\hspace{10pt}\,\\
Looking at the results for a small number of labeled training data, it can be seen for all data sets that the presented implementation of VAE forms a better predictive measure based on the extracted features in the range up to at least $\SI{2}{\percent}$. This supports the research hypothesis that autoencoder-based representation learning methods for extracting features from heterogeneous multivariate time series are particularly suitable for very small amounts of existing labeled training data in the presence of a sufficiently large amount of unlabeled training data at the same time.
    
\section{Conclusion}
In this work, an unsupervised autoencoder-based feature extractor for estimation tasks on heterogeneous, multivariate time series of mechatronic data was presented and validated on three data sets from the scientific community. Even though the prediction quality based on the entire training data does not approach current state-of-the-art feature extractors like \textit{Rocket}, an increased quality for the case of a low amount of labeled training data with a high availability of unlabeled data could be shown.\\
The presented results were obtained using only processed (reference) data sets. Thus, a consequential step towards automated estimation methods for the condition monitoring in comparable applications is the adaptation of the presented method for non-preprocessed raw data. Here, for example, the method of the \textit{denoising autoencoder} may be considered as a possible extension for noisy or even corrupted input data.
\vspace{-0.3cm}
\paragraph*{Publication notice}
This is a pre-print version of the paper in German language submitted to \textit{VDI Mechatronic Tagung 2021}, which has been published in the conference proceedings.

\clearpage
\onecolumn
\section*{Appendix}
\begin{table*}[hbt!]
    \centering
    \caption{Test results of the classification tasks with all three data sets. The mean and standard deviation of the accuracy for different proportions of labeled training data are given in each case.}
    \label{tbl:results_klass}
    \scriptsize
    \vspace{4mm}
        \begin{tabular}{C{4.5cm}cccc}
        \hline 
        {\% of labeled training samples ($n_{\mathrm{train}}$)} &
          {VAE + SVM} &
          {PCA + Ridge} &
          {tsfresh + Ridge} &
          {Rocket + Ridge} \\ \hline
        \multicolumn{5}{c}{Rolling bearing damages data set} \\ \hline
        $1.0\%\;(14)$ &  $\mathbf{{.541\pm.023}}$ &  $.424\pm.015$ &  $.484\pm.026$ &  \cmmnt{$.258\pm.000$ &}  $.424\pm.009$ \\
     $2.0\%\;(28)$ &  $\mathbf{{.545\pm.021}}$ &  $.383\pm.029$ &  $.496\pm.027$ &              $.426\pm.008$ \\
     $5.0\%\;(72)$ &             $.618\pm.023$ &  $.529\pm.029$ &  $.536\pm.027$ &   $\mathbf{{.636\pm.032}}$ \\
   $10.0\%\;(144)$ &             $.724\pm.026$ &  $.443\pm.028$ &  $.477\pm.020$ &   $\mathbf{{.817\pm.020}}$ \\
   $20.0\%\;(288)$ &             $.735\pm.012$ &  $.372\pm.020$ &  $.490\pm.017$ &   $\mathbf{{.862\pm.019}}$ \\
   $50.0\%\;(720)$ &             $.781\pm.015$ &  $.438\pm.017$ &  $.417\pm.001$ &   $\mathbf{{.827\pm.016}}$ \\
 $100.0\%\;(1440)$ &             $.768\pm.014$ &  $.512\pm.000$ &  $.443\pm.000$ &   $\mathbf{{.895\pm.006}}$  \\ \hline
        \multicolumn{5}{c}{Stepper motors data set} \\ \hline
        $0.1\%\;(70)$ &  $\mathbf{{.677\pm.013}}$ &  $.454\pm.016$ &  $.578\pm.013$ &  \cmmnt{$.639\pm.010$ &}   $.510\pm.014$ \\
     $0.2\%\;(140)$ &  $\mathbf{{.730\pm.025}}$ &  $.479\pm.014$ &  $.607\pm.012$ &              $.616\pm.016$ \\
     $0.5\%\;(350)$ &  $\mathbf{{.765\pm.010}}$ &  $.474\pm.015$ &  $.594\pm.010$ &              $.700\pm.033$ \\
     $1.0\%\;(701)$ &  $\mathbf{{.790\pm.015}}$ &  $.499\pm.017$ &  $.658\pm.017$ &               $.741\pm.036$ \\
    $2.0\%\;(1403)$ &  $\mathbf{{.808\pm.017}}$ &  $.502\pm.023$ &  $.663\pm.012$ &               $.774\pm.032$ \\
    $5.0\%\;(3507)$ &             $.820\pm.014$ &  $.512\pm.015$ &  $.677\pm.011$ &    $\mathbf{{.821\pm.011}}$ \\
   $10.0\%\;(7015)$ &             $.825\pm.013$ &  $.512\pm.012$ &  $.672\pm.013$ &    $\mathbf{{.866\pm.012}}$ \\
  $20.0\%\;(14030)$ &             $.826\pm.011$ &  $.515\pm.010$ &  $.679\pm.010$ &    $\mathbf{{.883\pm.013}}$ \\
  $50.0\%\;(35076)$ &             $.823\pm.011$ &  $.517\pm.012$ &  $.681\pm.011$ &    $\mathbf{{.902\pm.013}}$ \\
 $100.0\%\;(70152)$ &             $.809\pm.010$ &  $.516\pm.010$ &  $.680\pm.010$ &    $\mathbf{{.906\pm.012}}$ \\
          \hline
        \multicolumn{5}{c}{Hydraulic system dataset} \\ \hline
        $1.0\%\;(15)$ &  $\mathbf{{.972\pm.006}}$ &             $.906\pm.018$ &  $.886\pm.021$ &  \cmmnt{$.684\pm.000$ &}  $.924\pm.012$ \\
     $2.0\%\;(30)$ &  $\mathbf{{.974\pm.003}}$ &             $.905\pm.025$ &  $.948\pm.007$ &              $.947\pm.011$ \\
     $5.0\%\;(77)$ &             $.986\pm.001$ &             $.980\pm.004$ &  $.980\pm.007$ &    $\mathbf{{.992\pm.001}}$ \\
   $10.0\%\;(154)$ &             $.984\pm.001$ &             $.994\pm.001$ &  $.990\pm.001$ &    $\mathbf{{.996\pm.000}}$ \\
   $20.0\%\;(308)$ &             $.991\pm.002$ &             $.994\pm.001$ &  $.992\pm.001$ &    $\mathbf{{.997\pm.001}}$ \\
   $50.0\%\;(772)$ &             $.995\pm.000$ &             $.996\pm.001$ &  $.996\pm.001$ &    $\mathbf{{.997\pm.001}}$ \\
 $100.0\%\;(1544)$ &             $.997\pm.000$ &             $.997\pm.000$ &  $.994\pm.000$ &    $\mathbf{{.998\pm.000}}$ \\ \hline
    \end{tabular}
\end{table*}
\begin{table*}[hbt!]
    \centering
    \caption{Test results for the regression task with the hydraulic system data set. The mean value and standard deviation of the normalized RMSE for the prediction of the pressure in the hydraulic accumulator for different proportions of labeled training data are given in each case.}
    \label{tbl:results_regr}
    \scriptsize
    \vspace{4mm}
        \begin{tabular}{C{4.5cm}cccc}
        \hline 
        {\% of labeled training samples ($n_{\mathrm{train}}$)} &
          {VAE + SVR} &
          {PCA + Ridge} &
          {tsfresh + Ridge} &
          {Rocket + Ridge} \\ \hline
        $1.0\%\;(15)$ &  $\mathbf{.814\pm.058}$ &  ${1.674\pm.350}$ &             $1.026\pm.013$ &  $.923\pm.033$ \\
     $2.0\%\;(30)$ &  $\mathbf{.669\pm.018}$ &              $.885\pm.107$ &  ${1.002\pm.019}$ &  $.856\pm.037$ \\
     $5.0\%\;(77)$ &  $\mathbf{.565\pm.033}$ &              $.595\pm.123$ &  ${3.429\pm.350}$ &  $.633\pm.022$ \\
   $10.0\%\;(154)$ &  $.433\pm.009$ &              $\mathbf{.405\pm.091}$ &  ${1.986\pm.350}$ &  $.541\pm.085$ \\
   $20.0\%\;(308)$ &  $\mathbf{.321\pm.014}$ &              $.458\pm.129$ &  ${2.138\pm.350}$ &  $.367\pm.023$ \\
   $50.0\%\;(772)$ &  $\mathbf{.217\pm.004}$ &              $.237\pm.009$ &   ${.915\pm.266}$ &  $.864\pm.350$ \\
 $100.0\%\;(1544)$ &  $\mathbf{.169\pm.001}$ &              $.224\pm.000$ &  ${1.217\pm.000}$ &  $.212\pm.008$ \\ \hline
    \end{tabular}
\end{table*}
\clearpage  
\normalsize


\end{document}